\documentclass{article}

\usepackage{arxiv}

\usepackage[utf8]{inputenc} % allow utf-8 input
\usepackage[T1]{fontenc}    % use 8-bit T1 fonts
\usepackage{hyperref}       % hyperlinks
\usepackage{url}            % simple URL typesetting
\usepackage{booktabs}       % professional-quality tables
\usepackage{amsfonts}       % blackboard math symbols
\usepackage{nicefrac}       % compact symbols for 1/2, etc.
\usepackage{microtype}      % microtypography
\usepackage{subcaption}
\usepackage{multirow}
\usepackage{multicol}
\usepackage{enumitem}
\usepackage{lipsum}
\usepackage{fancyhdr}       % header
\usepackage{graphicx}       % graphics
\graphicspath{{media/}}     % organize your images and other figures under media/ folder

\title{Judging It, Washing It: Scoring and Greenwashing Corporate Climate Disclosures using Large Language Models}

\author{
Marianne Chuang$^1$\thanks{\hspace{2mm}Denotes co-first authorship, ordered randomly. Co-first authors will prioritize their names on their resumes/websites.} $\quad$ Gabriel Chuang$^{2*}$ $\quad$ Cheryl Chuang$^{1*}$ $\quad$ John Chuang$^{3}$ \\\\
$^1$ UC Santa Cruz \quad $^2$ Columbia University \quad $^3$ UC Berkeley
}

\begin{document}
\maketitle

\begin{abstract}
We study the use of large language models (LLMs) to both evaluate and greenwash corporate climate disclosures. First, we investigate the use of the LLM-as-a-Judge (LLMJ) methodology for scoring company-submitted reports on emissions reduction targets and progress. Second, we probe the behavior of an LLM when it is prompted to greenwash a response subject to accuracy and length constraints. Finally, we test the robustness of the LLMJ methodology against responses that may be greenwashed using an LLM. We find that two LLMJ scoring systems, \textit{numerical rating} and \textit{pairwise comparison}, are effective in distinguishing high-performing companies from others, with the pairwise comparison system showing greater robustness against LLM-greenwashed responses.
\end{abstract}

% keywords can be removed
% \keywords{First keyword \and Second keyword \and More}

\section{Introduction}
In the face of global climate change, corporations around the world are undertaking climate action plans, setting targets and making progress to reduce the carbon emissions of their operations and their supply chains. These actions are important not just for climate change mitigation and regulatory compliance, but also for the long-term sustainability and resilience of their businesses.

Corporate climate disclosures are a critical component of corporate climate action. They report information on their climate-related risks, emission reduction strategies and targets, and offer progress updates on a regular basis. Through these disclosures, corporations can provide transparency and accountability to their stakeholders, including investors, regulators, and consumers. 
Various reporting frameworks have been widely used, including CDP (formerly Carbon Disclosure Project), TCFD (Task Force on Climate-related Financial Disclosures), CSRD (Corporate Sustainability Reporting Directive),  % GRI (Global Reporting Initiative), SASB (Sustainability Accounting Standards Board),
and efforts are underway to harmonize and standardize them. 
The number of reporting companies is growing rapidly. For example, the number of companies voluntarily disclosing to CDP increased from 2,600 in 2018 to 23,000 in 2023. The European Union is expecting 50,000 companies to report to CSRD in 2025.

These disclosure reports can be comprehensive in scope, covering governance, strategy, risk management, metrics and targets, and the vast amounts of unstructured textual data make analysis a challenging task. Natural Language Processing (NLP) methods, including Large Language Models (LLMs), are emerging as important tools for analysts to extract key metrics, track progress, assess risks, and compare companies against their peers. 

Unfortunately, greenwashing in corporate climate disclosures is a real and growing problem. Greenwashing occurs when companies mislead their stakeholders into thinking that they are more environmentally responsible than they really are. By using vague, inaccurate, or noncommittal language, or by making unverifiable claims, companies can greenwash their disclosure reports, placing more pressure on stakeholders to critically assess their climate claims.

In this paper, we study the use of LLMs by analysts to evaluate corporate climate disclosures, as well as the use of LLMs by companies to enhance their disclosures, with or without the intention to greenwash. 

First, we investigate the use of the LLM-as-a-Judge (LLMJ) methodology \cite{zheng2023judging} to score the responses submitted by companies on their emission reduction targets and progress. Using a data set of 1,410 reports submitted to the CDP, we tested different variants of LLMJ to compare their performance. We find that two LLMJ scoring systems, \textit{reference-guided numerical rating} and \textit{pairwise comparison}, are effective in differentiating high-performing companies from others. We also find the use of various LLM techniques, such as in-context learning, indicative scales, and chain-of-thought prompting (via explanation requirements), can provide performance improvements in different contexts.

Second, we performed a series of experiments to learn how an LLM can be used by companies to improve their responses, and to test the robustness of the LLMJ methodology against responses that may be greenwashed using an LLM. We find that, when unconstrained, an LLM is great at greenwashing, especially for low-rated responses. It can fabricate lengthy, plausible-sounding content with little connection to the original, and it can turn proposed plans into completed actions, or planned targets into achieved targets. However, when accuracy requirements are put in place, the LLM will shift its focus to improve the clarity of the writing, generate longer responses to elaborate on a company’s reported plans and progress, or add aspirational language that are not verifiable nor tied to emissions targets or progress. In this latter case, where hallucinated, factually false content is not present in the responses, the LLMJ, particularly the pairwise comparison scoring system, is able to retain its robustness against LLM-enhanced responses.

\section{Related Work}
There is a quickly growing body of literature on the use of Natural Language Processing (NLP) and machine learning methods to contribute to tackling climate change \cite{stede2021climate,rolnick2022tackling}. They include efforts to detect, analyze, and fact-check environmental claims and stances \cite{leippold2023environmental,luo2020detecting,coan2021computer,piskorski2022exploring,gehring2023analyzing,diggelmann2020climate,stammbach2022environmental,morio2023nlp}, identify topics and trends over time \cite{yim2023meticulously,brie2024mandatory}, improve the performance of conversational AI agents with regards to climate change related information \cite{webersinke2021climatebert,vaghefi2023chatclimate,bulian2023assessing}, and tools to support climate policymaking \cite{callaghan2021machine,planas2022beyond}.

There is also a number of recent works that employ LLMs to analyze environmental assessment reports, corporate sustainability reports, and corporate climate disclosure documents. The LLMs have proven themselves to be very versatile, capable of sifting through lengthy documents to detect and extract specific items of interest, such as emission reduction targets \cite{schimanski2023climatebert,wrzalik2024netzerofacts} and sustainable development goals \cite{garigliotti2024sdg}. Furthermore, the LLMs can also be used to analyze entire reports to generate overall assessments of a company’s performance or transition plans \cite{ni2023chatreport,colesanti2024combining}.

While not yet reported in the wild, we can expect that LLMs will soon be recruited for greenwashing \cite{moodaley2023greenwashing}. For example, researchers recently used an LLM to generate fictional sustainability reports, demonstrating both the potential and current limitations of the technology \cite{de2024will}. Conversely, researchers have shown that LLMs can be effective in detecting cheap talk, cherry picking, and exaggerations \cite{bingler2022cheap,luo2024unmasking}. 

The LLM-as-a-Judge (LLMJ) method has recently emerged as a powerful tool to perform evaluation tasks across a wide range of domains in a scalable manner \cite{zheng2023judging}. LLM judges can flexibly adjust their evaluation criteria, and generatively produce evaluation outputs, based on the specific contexts of the task. While the LLMJ method inherits a number of limitations from LLMs (e.g., hallucinations and domain-specific knowledge gaps), and exhibits vulnerabilities to biases (e.g., position bias and verbosity bias), their negative effects can be mitigated with prompt engineering and other measures. While the LLMJ method was originally proposed for evaluating chatbot responses, it has since been applied to domains including law \cite{yue2023disc}, finance \cite{son2024krx}, medicine \cite{xie2024doclens}, and education \cite{chiang2024large,wang2024automated}. However, to the best of our knowledge, this paper is the first to use the LLMJ method in the climate and sustainability domain.

\section{Data}

CDP was established as the `Carbon Disclosure Project' in 2000, and collects voluntary climate disclosures via their Climate Change Questionnaire from companies on an annual basis. Since 2013, CDP also compiles an annual “A-List” of companies that meet their criteria to be considered leaders on environmental transparency and action. The annual questionnaire includes more than a dozen sections, covering a wide range of topics such as governance, risks and opportunities, business strategy, verification, carbon pricing, and engagement. In this study, we focus on the first two questions in the "Targets and Performance" section: 

\begin{itemize}[nosep]
    \item \textbf{4.1a}: Provide details of your absolute emissions target(s) and progress made against those target(s).
    \item \textbf{4.1b}: Provide details of your emissions intensity target(s) and progress made against those target(s).
\end{itemize}

“Absolute emissions” refers to the total quantity of emissions (i.e., tons of carbon), whereas “emissions intensity” refers to an amount that is relative to the size of the company. Each reflects a different aspect of a company’s targets and progress, and both are important for a complete assessment.

We use the CDP dataset from 2022, which consists of 8385 global companies, 2398 of which are from Europe. We focus on the 1416 European companies that submitted a response to Question 4.1a and/or 4.1b, of which 147 made the ``A-List''. 

\section{LLM-as-a-Judge (LLMJ) for Climate Disclosures}
\label{section:LLMJ}

\begin{figure*}[h]
  \centering
  \includegraphics[width=\linewidth]{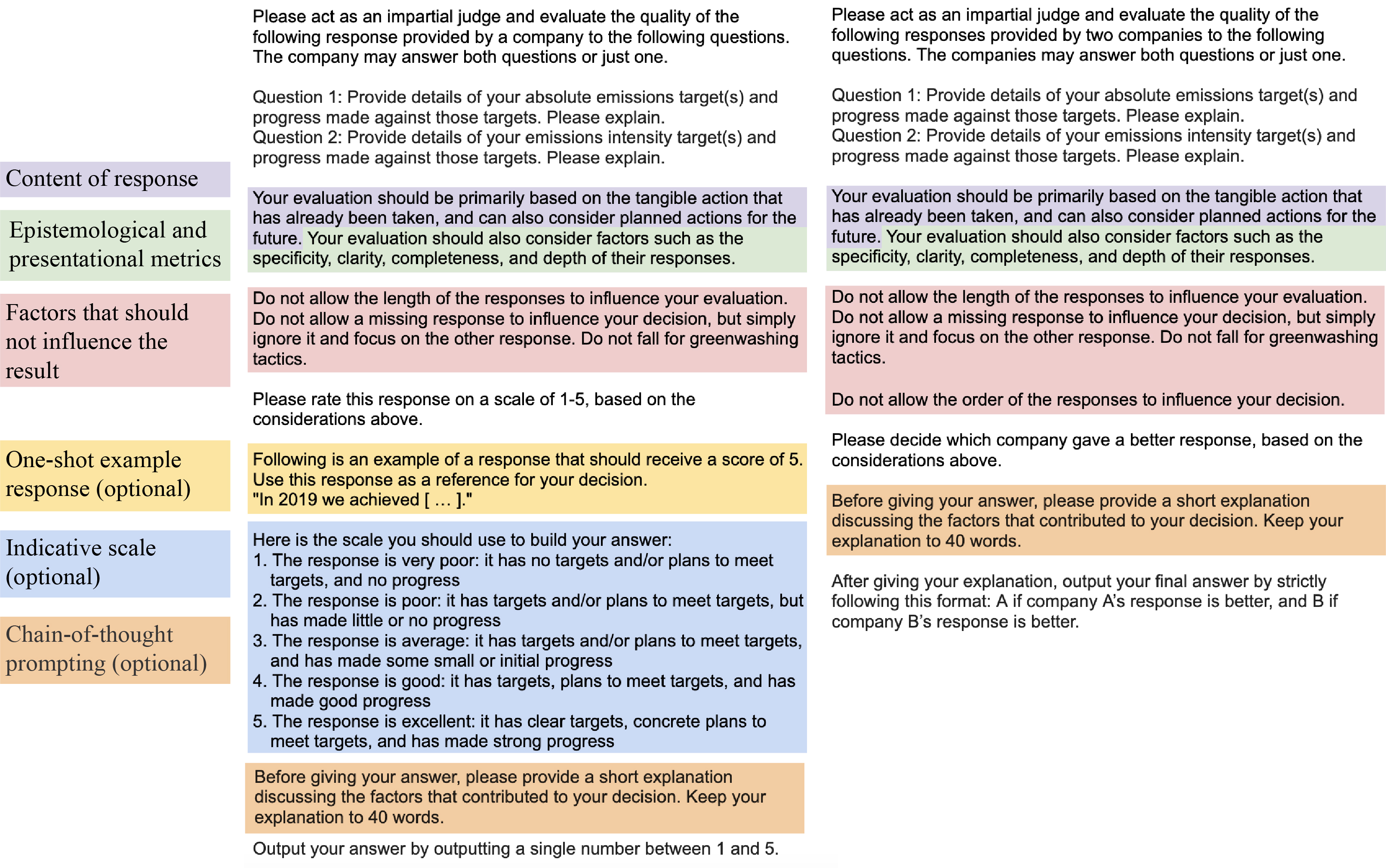}
  \caption{Baseline LLM-as-a-Judge prompts for  {\itshape numerical rating} (left) and {\itshape pairwise comparison} (right).}
  %\Description{Text of the baseline prompt used in our study.}
  \label{fig:prompt}
\end{figure*}

The premise of the LLM-as-a-Judge technique  \cite{zheng2023judging}, is to use a particular prompting setup to guide an LLM in giving a score to a piece of text. In this work, we evaluate two different scoring systems: {\itshape numerical rating} (e.g., ``rate this response on a scale of 1 to 5''), and {\itshape pairwise comparison} (e.g., ``which of these two responses is better?''). A labeled sample prompt for each system is shown in Figure~\ref{fig:prompt}.

In both scoring types, we follow \cite{bulian2023assessing} in asking the LLM to consider accuracy, specificity, and completeness (``epistemological metrics'') and clarity (a ``presentational metric''), in addition to the actual content of the response. We also specify factors that the LLM should {\itshape not} consider, such as the raw length of the response or irrelevant information.\footnote{We use OpenAI’s GPT-4o-mini-2024-07-18 for our experiments, sampling with temperature parameter \textit{t} = 0. Our code is publicly available at https://github.com/mariannechuang/llm-corp-disclosure.}
% All prompt templates are also listed in Appendix A. 

\subsection{Numerical Rating}

In the numerical rating scoring system, we ask the LLM to give the response a numerical score from 1 to 5. Because language models output tokens non-deterministically, we compute a weighted average, weighting each potential response (1 through 5) by the probability of outputting that response.\footnote{The OpenAI API allows users to request a distribution over next token predictions, rather than the single sampled next token. We find that using the weighted average rather than simply the sampled output token improves results. For details, see Appendix~\ref{appendix:logprobs}.}

\subsection{Pairwise Comparison}

In the pairwise comparison scoring system, we score a response by individually comparing it to $k$ other uniformly selected responses, asking the LLM to evaluate which response is “better” and ranking the response overall in terms of its “expected win rate,” out of 100\%. For example, a response which is rated “better” in 15 comparisons and “worse” in 5 would receive a score of 75 out of 100.

Again, because language model outputs are non-deterministic, each pairwise comparison yields a probability of each outcome, rather than a direct binary outcome.% (for example, the LLM may return response A with probability 0.9 and response B with probability 0.1). We simply sum up the total “win probability” from each comparison and divide by the number of comparisons to get a final score.
We simply compute the expected "win percent" over all comparisons. Notably, pairwise comparison is much more computationally expensive than numerical rating, because it requires $k$ queries per response.

%For pairwise comparison, we test one additional variable: asking the LLM to explain its reasoning before giving its choice, which is known as chain-of-thought prompting.

\subsection{Variable Prompt Sections}
We test three additional variables: 
\begin{itemize}[nosep]
    \item Providing reference responses (i.e., in-context learning), which we test for numerical rating; 
    \item Using an indicative scale, which we test for numerical rating; 
    \item Chain-of-thought prompting (i.e., asking the LLM to explain its answer), which we test for both  numerical rating and pairwise comparison. 
\end{itemize}

\subsubsection{In-Context Learning} It is known that LLMs are few-shot learners \cite{brown2020language}: that is, they can perform tasks given only a small number of examples and without additional fine-tuning or gradient updates. On the other hand, modern LLMs like GPT-4 have been trained on text corpora of sufficiently massive scale that they are also often able to perform tasks given only instructions, without any examples provided. Thus, for the specific domain of climate disclosures, it is not immediately obvious whether reference examples are needed or whether the pre-trained “knowledge base” of the LLM is sufficient. 
To explore this, we test three configurations, providing the LLM with:
\begin{itemize}[noitemsep,topsep=2pt]
  \item No example responses (zero-shot learning);
  \item One example response, which should receive a score of 5 (one-shot learning); and
  \item Two example responses that should receive scores of 3 and 5 (few-shot learning). 
\end{itemize}
We manually select representative examples for each score from our dataset. 

\subsubsection{Indicative Scale} Numerical rating systems often feature indicative scales, which describe what each numerical value represents. Examples include the Likert scale (from “Strongly Disagree” to “Agree”) and the pain scale (which uses images to indicate levels of pain from 1 to 10). Some sources suggest that an indicative scale can help LLM-as-a-Judge systems \cite{Roucher2025using}. We construct a scale based on the reported targets, plans, and progress. The scale is highlighted in blue in Figure~\ref{fig:prompt}.

\subsubsection{Chain-of-Thought Prompting} Chain-of-thought prompting is a technique that asks the language model to perform intermediate reasoning steps before coming up with a final answer. It has been shown to substantially improve performance on reasoning-based tasks such as arithmetic and symbolic reasoning tasks \cite{wei2022chain}. To date, there is no work in the literature on using chain-of-thought for LLMJ tasks; so, we test our scoring systems both with and without chain-of-thought prompts. We ask the LLM to produce a short explanation before making its final decision for each choice. To reduce computational burden, we limit the explanations to 40 words. 

\subsection{Evaluation}

To evaluate the scoring systems, we focus on their ability to distinguish high-performing companies from others, as determined by the CDP’s “A-List”. In particular, we compare the distribution of scores received by A-List companies against the distribution of scores received by non-A-List companies. %In 2022, there were 147 A-List and 1269 non-A-List European companies that answered at least one of Questions 4.1(a) and 4.1(b). 

For the numerical rating system, we simply compute the weighted score (from 1-5) for all 1,416 companies. For the more costly pairwise comparison system, we evaluated all 147 A-List companies and randomly sampled 147 non-A-List companies, and performed $k=24$ comparisons against a random sample of the entire set. %To reduce randomness, both the set of 147 non-A-List companies and the 24 “comparison” companies were sampled in a stratified way based on their numerical rating scores. %In order to make an apples-to-apples comparison, we bucket the numerical rating scores into 25 buckets of equal size from 1 to 5. 
To make an apples-to-apples comparison, we bucket the numerical rating scores into 25 bins of equal width from 1 to 5. 

\interfootnotelinepenalty=10000

In each case, we measure the distance between the two distributions (A-List and non-A-List) using three standard distance measures: Total Variation Distance (TVD), the Kolmogorov-Smirnov (KS) statistic, and the normalized Earth Mover Distance (EMD). The TVD captures the overall overlap of the probability mass of the two distributions. The KS statistic captures the maximum cumulative difference, loosely corresponding to the separation of the best threshold predictor if both distributions occurred at equal base rates. The EMD captures the distance between the non-overlapping probability mass of the distributions, relative to the overall range of possible outcomes.\footnote{We normalize the EMD metric because pairwise comparison yields scores from 0 to 100 while numerical rating yields scores from 1 to 5.}%All three are lower for distributions that are more similar. 

\subsection{Results}

\begin{table*}[t]
  \caption{Separation between score distributions of A-List and non-A-List company responses, as measured by Total Variation Distance, Kolmogorov-Smirnov statistic, and normalized Earth Mover Distance metrics. %Higher separation means the configuration is better able to distinguish high-performing company responses. The EMD metric is the most discriminative metric for comparing the different scoring systems and prompt configurations. pairwise comparison offers significantly higher EMD than numerical rating. Giving reference example(s) notably improves the separation, using chain-of-thought prompting yields smaller improvements, while using indicative scale does not offer meaningful improvements.
  }
  \label{tab:results}
  \centering 
  \begin{tabular}{llccc}
    \toprule
    Scoring System & Prompt Configuration & TVD & KS & EMD\\
    \midrule
    Numerical rating & zero-shot & 0.4000 & 0.4169 & 0.1188\\
    & zero-shot, indicative scale & 0.2738 & 0.3859 & 0.1050\\
    & one-shot & 0.4423 & 0.4413 & 0.1799\\
    & one-shot, indicative scale & 0.4422 & 0.4413 & 0.1842\\
    & one-shot, chain-of-thought & 0.3945 & 0.3940 & 0.1346 \\
    & two-shot & 0.4612 & 0.4600 & 0.1715\\
    & two-shot, indicative scale & 0.4401 & 0.4431 & 0.1776\\
    \midrule
    Pairwise comparison & no chain-of-thought & 0.4724 & 0.4432 & 0.3099\\
    & chain-of-thought prompting & 0.4855 & 0.4508 & 0.3145\\
  \bottomrule
\end{tabular}
\end{table*}

\begin{figure*}
    \centering
    % Left figure (2/3 width) with two subfigures
    \begin{minipage}[t]{0.64\textwidth}
        \vspace{0pt}
        \centering
        \begin{subfigure}{0.49\textwidth} 
            \centering
            \includegraphics[width=\linewidth]{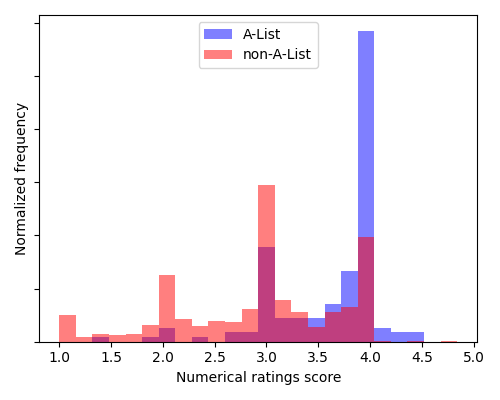}
            \caption{Scores under numerical rating with one-shot reference}
        \end{subfigure}
        \hfill
        \begin{subfigure}{0.49\textwidth}
            \centering 
            \includegraphics[width=\linewidth]{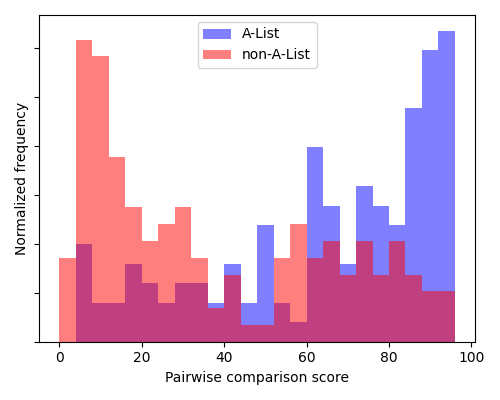}
            \caption{Scores under pairwise comparison without chain-of-thought prompting.}
            \label{fig:pairwise_distribution}
        \end{subfigure}
        \caption{Distribution of scores for A-List and non-A-List companies from numerical rating (left) and pairwise comparison (right).}
        \label{fig:score_distributions}
    \end{minipage}
    \hfill
    % Right figure (1/3 width)
    \begin{minipage}[t]{0.32\textwidth}
        \vspace{0pt}
        \centering
        \includegraphics[width=\linewidth]{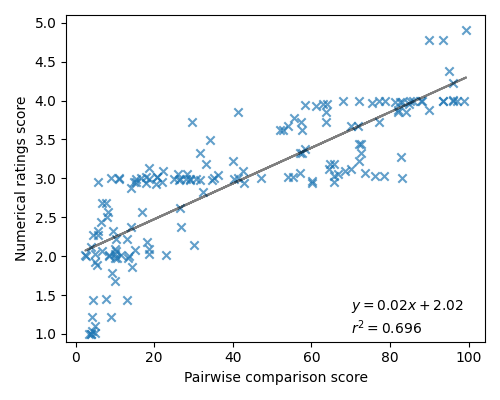}
        \caption{The two scoring systems produce consistent results: there is high correlation between the two scores
        ($r^2 = 0.70$).}
        \label{fig:pairwise_ratings_correlation}
    \end{minipage}
    %\caption{Overall Figure Caption}
\end{figure*}

The TVD, KS, and EMD values for each configuration, measuring the separation between A-List and non-A-List scores, are shown in Table ~\ref{tab:results}. We also show the overall distribution of scores for two of the configurations %(Numerical rating with one-shot reference and pairwise comparison with chain-of-thought prompting) 
in Figure~\ref{fig:score_distributions}. The distributions of scores for other configurations can be found in Appendix~\ref{appendix:all_hists}. We make the following observations:

\textbf{Both scoring systems separate high-performing and low-performing responses fairly well, with pairwise comparison outperforming numerical rating.} The overlap between the distributions is relatively small. In particular, we note that we do not expect to achieve anywhere near full separation of the two distributions: we use A-List status as only a rough proxy for the quality of the company’s response to these two specific questions; in reality, A-List status is determined based on an elaborate methodology to score the responses to these and dozens of other questions in the questionnaire \cite{CDP2022Scoring}.

\textbf{The two scoring systems create very differently-shaped score distributions.} The numerical rating system results in mostly near-integral scores (1 through 5) - that is, the LLM nearly always samples its answer from a distribution where an overwhelming proportion of the weight is on a single answer. %In particular, across all experiments, {\color{red} x\%} of responses had at least 95\% of the next-token distribution on a single token. 
On the other hand, the pairwise comparison scores are much more spread out: as $k$ grows large, we expect the distribution of all scores to converge to uniform over the $k$ bins.\footnote{Note that simply summing the red and blue bars in the histogram in Figure~\ref{fig:pairwise_distribution} will not create the uniform distribution because there are much fewer A-List companies than non-A-List companies, i.e., summing will oversample A-List companies.}

\textbf{The two scoring systems produce consistent results.} As shown in Figure~\ref{fig:pairwise_ratings_correlation}, the scores given by the two systems are highly correlated, with $r^2 = 0.70$. 

\textbf{Using at least one reference example is helpful.} There is a clear increase in separation when going from zero-shot to one-shot prompting. However, going from one reference example to two does not clearly show any additional improvement. \textbf{Using an indicative scale does not seem to improve separation,} but does change the distribution of scores. Our particular choice of indicative scale shifted responses away from scores near 3 and towards scores near 2 and 4 (see Figure~\ref{fig:all_hists_ratings}(b) versus ~\ref{fig:all_hists_ratings}(e)); this suggests that one can roughly tune the score distribution by carefully choosing the scale. For pairwise comparison scoring, \textbf{chain-of-thought prompting is moderately helpful}, but seems detrimental for numerical rating. %resulting in small increases in separation at the cost of substantially increased inference-time computation. 

Overall, these results are very promising. Given that our labels (A-List versus non-A-List) are very coarse, and that the responses we are scoring are in reality only one part of the consideration for A-List status, the fact that both scoring systems can capture a very substantial amount of signal is remarkable. Given the choice, it seems that the pairwise comparison system produces better results. However, the numerical rating system has its own advantages, e.g., it is much less computationally expensive, and it can avoid any potential moral or legal concerns regarding the comparison of companies' responses against each other. In addition, the pairwise comparison system may be subject to inflation/deflation over time, since the median score of 50 will track with the quality of the median response over time. Whether this quality is desirable or undesirable will likely depend on the particular goal of the assessor.

\section{Greenwashing with LLMs, and LLMJ Robustness against Greenwashing}

%In Section ~\ref{section:LLMJ}, we have discussed the role of LLMs in evaluating human-written climate disclosure responses, and presented two LLM-as-a-Judge scoring systems that perform well in separating high-performing and low-performing companies’ responses. 
Next, we investigate the intersection of LLM-based systems and greenwashing. In particular, we highlight two areas of overlap:
\begin{itemize}[topsep=2px, noitemsep]
  \item First, LLMs can be used to \emph{perform} greenwashing. Given the natural fit between LLMs and text-based tasks like greenwashing, the relevant question is not whether LLMs can perform greenwashing but how effectively they can do so. In addition, there may be various constraints imposed: for example, companies may want to make only surface-level changes to their responses, rather than add verifiably false information or clearly unachievable targets. More benignly, there may be limits on response length, which an LLM-modified response would need to adhere to. 
  \item Second, it is important to know whether the LLM-as-a-Judge systems discussed above will be susceptible to being “fooled” by greenwashed responses.
\end{itemize}

To begin to answer these questions, we generate a set of LLM-greenwashed responses, and then score these responses using the LLMJ systems from Section ~\ref{section:LLMJ}. By comparing the before-and-after greenwashing scores, we can evaluate the effectiveness of LLM-based greenwashing under various constraints on accuracy and length. In addition, by comparing the separation of greenwashed low-performing responses against non-greenwashed high-performing responses, we can evaluate the robustness of our LLM-based scoring systems against greenwashing.

\subsection{Greenwashing with LLMs}

We consider three types of greenwashing:
\begin{enumerate}[noitemsep, topsep=2px]
  \item “Unconstrained,” imposing no limits on the changes made by the LLM;  
  \item “Fixed accuracy,” requiring that the LLM not describe fictitious actions or results; and 
  \item “Fixed accuracy and length,” requiring that the LLM also preserve the length of the original response (in addition to maintaining factual information). 
\end{enumerate}

\begin{figure*}
   \centering
   \includegraphics[width=\linewidth]{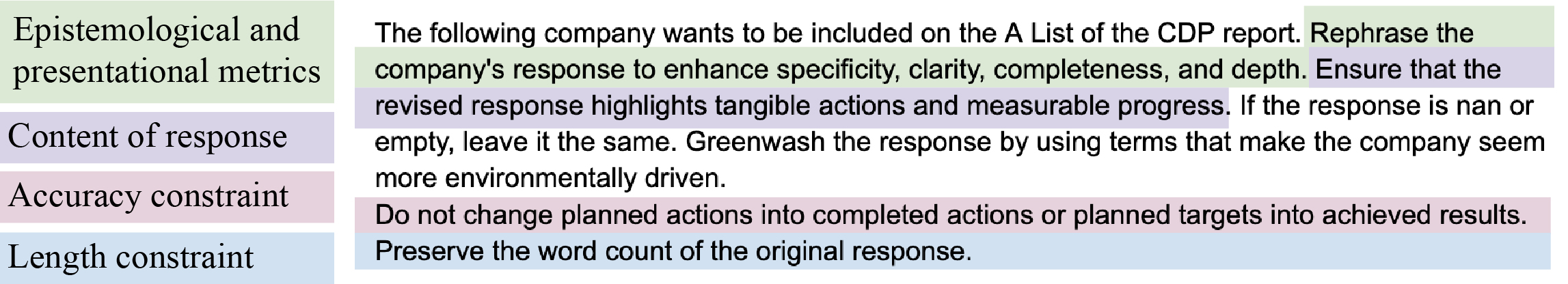}
   \caption{The prompt used to generate greenwashed responses. We generated responses with no constraints, accuracy constraints (red), and both accuracy (red) and length (blue) constraints.}
   \label{fig:greenwash_prompt}
\end{figure*}

\begin{figure*}
   \centering
   \includegraphics[width=\linewidth]{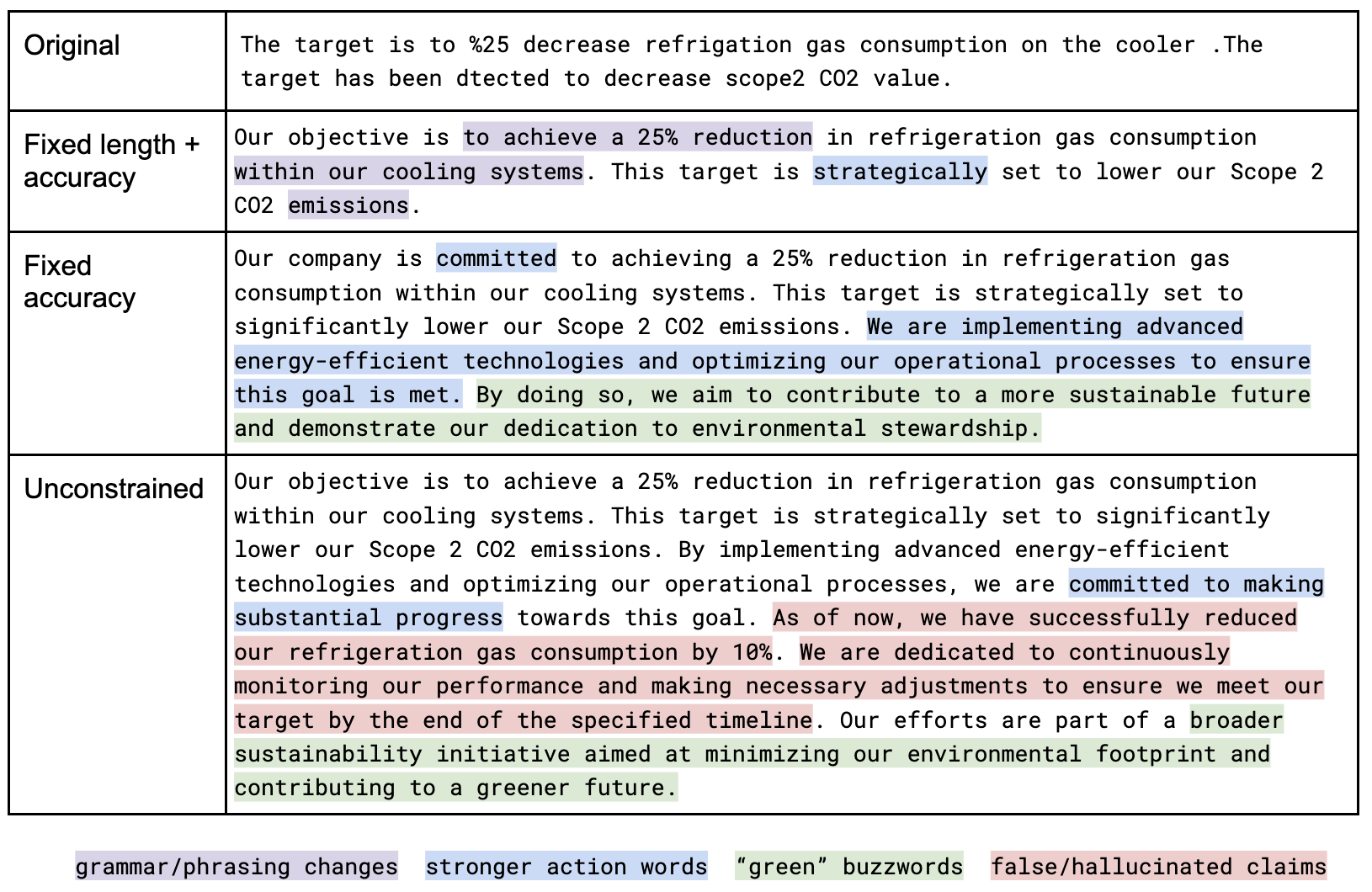}
   \caption{An example CDP response, along with LLM-greenwashed variations under three sets of constraints. Changes are loosely labeled by type.}
   \label{fig:greenwash_example}
\end{figure*}

We uniformly sample a set of 100 non-A-list companies from the CDP dataset, and prompt GPT-4o-mini to greenwash each of them, with the prompt shown in Figure~\ref{fig:greenwash_prompt}, 
resulting in three sets of greenwashed responses. Examples of the generated responses of each type are shown in Figure~\ref{fig:greenwash_example}.

We score each set of greenwashed responses using the two scoring systems from Section~\ref{section:LLMJ}: numerical rating (with one-shot reference), and pairwise comparison. The average scores of the original responses, in addition to the three variations of constrained greenwashing, are shown in Table~\ref{tab:greenwashed_scores}. %The full distribution of scores given by each scoring system to each set of greenwashed responses is shown in Appendix~\ref{appendix:TODO}.

\begin{table*}
  \caption{Changes in average LLM-as-a-Judge numerical and pairwise rating scores given to LLM-greenwashed responses, compared to original.}
  \label{tab:greenwashed_scores}
  \begin{tabular}{c|cccc}
    \toprule
    & \multirow{2}{*}{Original} & \multicolumn{3}{c}{Greenwashed} \\
    & & Fixed length \& accuracy & Fixed length & Unconstrained \\
    \midrule
    Average rating score & $2.963$ & 3.202 $(+0.239)$ &3.520 $(+0.557)$ & 3.591 $(+0.628)$\\ 
    Average pairwise score & 48.2 & 50.8 $(+2.6)$ & 58.7 $(+10.5)$ & 61.8 $(+13.6)$\\
    \midrule 
    EMD vs. A-List (rating) & 0.17 & 0.11 $(-0.06)$ & 0.03 $(-0.14)$ & 0.02 $(-0.15)$ \\ 
    EMD vs. A-List (pairwise) & 0.24 & 0.18 $(-0.06)$ & 0.11 $(-0.13)$ & 0.08 $(-0.16)$ \\
  \bottomrule
\end{tabular}
\end{table*}

We make the following observations: 

First, GPT-4o-mini is, unsurprisingly, quite capable at greenwashing, particularly when it is allowed to hallucinate plans, goals, actions taken, and so on, generating responses that score an average of 0.63 points higher on the 5-point numerical rating scale and 14 points higher on the 100-point pairwise comparison scale. On the numerical rating scale, 55\% of responses saw a score increase of at least half a point, and 17\% saw a score increase of 1 point or more. 

Qualitatively, \textbf{we observe the LLM making several different types of changes in its generated responses} (roughly ordered by amount of change):
\begin{enumerate}[noitemsep, topsep=2px]
    \item Grammar, spelling, and wording changes; These are particularly common among companies that are not based in English-speaking countries. 
    \item Using stronger, action-oriented language with the same meaning as the original (``strongly committed,'' ``intensely focused,'' etc.);
    \item Adding ``green buzzwords,''such as ``ensuring a greener future'' or ``environmental stewardship,'' which vaguely describe high-level ideals without mentioning specific plans, targets, or actions taken; 
    \item Adding (or alluding to) vague, unspecified plans to meet specific stated goals;
    \item Adding completely false/hallucinated information, mostly about targets met (for example, ``We reduced our Scope 2 emissions by 10\% over the last year.''). This only occurs in the unconstrained case. In particular, we often observe the LLM changing planned actions (in the original response) to achieved actions (in the modified one). 
\end{enumerate}

In the ``fixed accuracy'' case, we only observe changes 1-4 above, resulting in smaller score increases of +0.56 points in numerical rating and 11 points in pairwise comparison. In the ``fixed accuracy and length'' case, we only observe changes 1-2 above, and see score increases of +0.24 in numerical rating and +3 points in pairwise comparison. 

\begin{figure}
    \centering
    \includegraphics[width=0.49\linewidth]{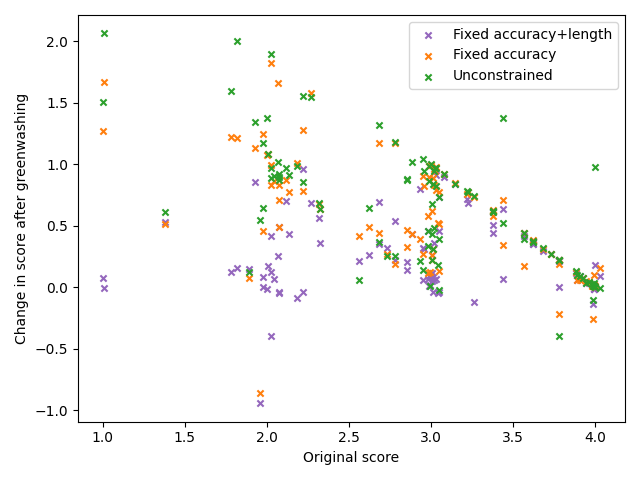}
    \caption{Score improvement plotted against original score, for the three sets of greenwashed responses. As expected, unconstrained greenwashing results in larger score increases than fixed-accuracy, and fixed-accuracy dominates fixed-accuracy-and-length. }
    \label{fig:score_change}
\end{figure}

\textbf{LLMs improve low-scoring responses more than high-scoring ones.} Figure~\ref{fig:score_change} shows the increase in score under each set of constraints, plotted against the original score. The responses with the largest score increases (around +2 points) were ones that began with original ratings around 1-2, with the ceiling on improvement decreasing linearly. Across a wide range of initial scores, most modified responses were capped at a final score of around 4 (even with unconstrained greenwashing).

\textbf{Even when given constraints, LLMs may not follow them when greenwashing.} Several of the “fixed accuracy” responses saw many fabricated claims, especially numerical or percentage-based emissions reduction amounts and made-up descriptions of actions taken. While worrying, this is unsurprising given the well-documented tendency of LLMs to hallucinate, and can be taken as a warning: such verifiable falsehoods would be caught by a careful human reader or auditor. As we will discuss below, this also complicates our discussion of the robustness of our LLM-as-a-Judge system. 

\textbf{Finally, the LLM sometimes replaces or obscures useful information with junk.} We observe multiple instances in the “fixed length and accuracy” set where the modified response replaces some useful information with generic platitudes. %For example, in one response, it substantially shortens an explanation of why the company reports absolute rather than relative figures (due to high variance in product energy requirements) with “Our commitment to these absolute targets underscores our dedication to tangible and measurable environmental improvements,” which adds nothing (and in fact decreases the response’s score by 2 points on the pairwise scale). 
In several instances, the LLM re-words clear descriptions into dense buzzword-heavy sentences, obscuring the practical information %Even when it preserves the factual information, the LLM often adds “fluff” sentences which obscure the truly informative content 
and adding ``fluff'' that makes the response harder (for a human) to read. 

\subsection{Robustness of LLMJ against Greenwashed Responses}

We make an exploratory discussion of the robustness of the LLM-as-a-Judge systems presented in Section~\ref{section:LLMJ} against greenwashed responses. 
When a greenwashed response receives a higher score, it can either be because the LLMJ system was fooled by surface-level changes (a failure of the scoring system) or because the greenwashed response introduced false information % that, if true, would genuinely increase the quality of the response 
(which would be unreasonable to expect the LLMJ to recognize). In the absence of expert-annotated labels of the greenwashed responses, it is difficult to definitively attribute score increases to one or the other. 

%Of course, when a greenwashed response receives a better score, it can be for one of two reasons (or a combination of both): either 
%\begin{enumerate}[noitemsep, topsep=2px]
%    \item The greenwasher only changed the style of the response and not the content, but the LLM-as-a-Judge scoring system was fooled by the greenwashing. This would be a case where the LLM-J was not robust against greenwashing. 
%    \item The greenwasher introduced false information that, if true, would genuinely increase the score of the response. In this case, one cannot expect the LLM-as-a-Judge system to be able to verify the claims, and so this is not a failure of the judge system. 
%\end{enumerate}

%We note that, in the absence of human- or expert-annotated labels of the greenwashed responses, it is impossible to definitively attribute score increases to one or the other. 
While we attempt to control for the latter case above by asking the LLM greenwasher to preserve the accuracy of its modified responses, it does not reliably follow these directions (as we note above). %Nevertheless, most of the “fixed accuracy and length” responses seem to preserve factual content, so 
We use “fixed accuracy and length” responses, which seem to hallucinate the least, as a rough proxy for surface-level changes.

\textbf{Overall, when (approximately) controlling for truthfulness of responses, the LLMJ system is quite robust.} When the greenwashed responses are constrained on length and accuracy, the mean score increases by only +2.6 out of 100 (for pairwise) and +0.24 out of 5 (for rating). 
Even with unconstrained greenwashing (i.e., allowing the LLM to make up actions and targets), very few responses saw their score increase by large amounts: only 7\% saw increases of above 40 points or higher (on pairwise) and 1.5 points or higher (on rating). This is fairly strong: this means, for example, that no responses were able to be greenwashed from receiving a 1/5 to a 4/5, or from a 30/100 to 80/100.

Given that the scoring systems are meant to help distinguish high-performing companies from low-performing ones (and conversely, greenwashing is meant to make low-performing companies appear to be high-performing), we examine the separation between the scores of the A-List companies and the greenwashed non-A-List companies. Given that the changes in raw scores are relatively small, we might expect the change in separation to be correspondingly small. On the contrary, we find that the normalized EMD drops dramatically, as shown in Table~\ref{tab:greenwashed_scores}%Figure~\ref{fig:EMD_change}
: that is, \textbf{relatively small absolute changes to the scores can make low-performing companies seem similar to high-performing ones.} %Furthermore, even though the distribution on score change are quite similar for the pairwise and rating systems, t

Because the EMD decreases at approximately the same absolute amount, \textbf{the pairwise system is more robust to greenwashing due to a higher baseline separation}. In the original score distributions (Figure~\ref{fig:score_distributions}), the pairwise scores are much more uniformly distributed, whereas the rating scores are concentrated among the central scores of 3 and 4; thus, the pairwise comparison system is more robust to a small amount of improvement on low-scoring responses. 

We further discuss the robustness of the two scoring systems, including comparing the distribution of score increases, examining the effect of length, and discussing correlation between the two systems, in Appendices~\ref{appendix:greenwashing_robustness} and~\ref{appendix:doubling_length}.

\section{Limitations and Future Work}

We limited our analysis to a slice of the CDP data corpus, focusing on corporate responses to a single set of questions (on emission reduction targets and progress) from a single year (2022) from a single geographic region (Europe). It would be valuable to test the generalizability of our findings across other questions (such as governance structures, risk management strategies, adoption of internal carbon prices), years, geographic regions, and even other reporting frameworks. At the same time, this points to opportunities for researchers to employ our LLMJ methodology to analyze company performance over time, and to extend it to evaluate progress at a sectoral or industry level.

Similarly, we ran our analysis on a single LLM (OpenAI's ChatGPT-4o-mini). Given that LLMs continue to evolve and improve at a rapid pace, it would be valuable to repeat the analysis on other state-of-the-art LLMs and future generations of LLMs, so that we can gather more data points on the performance of the LLMJ method and its various in-context learning, indicative scale, and chain-of-thought techniques against different language models. By using different LLMs to evaluate LLM-generated greenwashed responses, one can also test for self-enhancement biases in the LLMJ methodology in this context.

The CDP publishes the comprehensive scoring methodology that they use to evaluate a company's response to each individual question in their annual questionnaire. However, CDP only publishes an overall "A-List" of high performing companies, without a breakdown of how each company scores for each individual question. Therefore, our study can only use a company's membership on the "A-List" as an indirect signal for high performance in the "targets and progress" aspect of their disclosure. While A-List companies are generally high-performing with regards to emission reduction targets and progress, we expect there may be other companies that are equally high-performing in this regard to nonetheless fail to achieve A-List status due to other deficiencies in their disclosures. This may have led to a conservative underestimation of the reported LLMJ performance numbers.

Our LLMJ scoring prompts simply ask the LLM to not fall for greenwashing tactics, but do not include any explicit greenwashing detection mechanisms. At the same time, our LLM greenwashing experiment reveals distinct ways an LLM may perform greenwashing. Future work can close the loop and study how  incorporating insights into the taxonomy and patterns of greenwashing may improve the performance of LLMJ scoring systems.

\section{Conclusion}

Our study finds that the LLM-as-a-Judge methodology can perform consistent, unbiased, and rules-based evaluations of corporate climate disclosures, and it does so in a performant and scalable manner. Furthermore, it offers robustness against greenwashing LLMs, short of hallucinated, factually false content.

We focus on scoring disclosures on emission reduction targets and progress, which is arguably the most tangible and direct way that a company's climate action can be tracked and evaluated. Recognizing the fact that the claims made by the companies may not have been fact-checked by CDP, our analysis shows that the LLMJ methodology can effectively evaluate the claims when taken at face value.

Our experiment shows that a greenwashing LLM can readily turn planned actions into achieved actions, either when it is unconstrained, or when it ignores accuracy requirements imposed by the prompt. However, since the disclosure responses ultimately have to be signed off by company officers, we should not expect the burden to fall on the LLMJ to distinguish between reportedly achieved actions that are real versus hallucinated.

While the pairwise comparison scoring system outperformed numerical rating on the EMD metric, we must recognize that it incurs significantly higher computational costs (by a factor of $k$, the number of companies to compare against). Further, there is the practical issue of gaining access to responses from a representative set of companies, either from the current year, or from a previous reference year. This may be particularly challenging for individual companies, so an organization like CDP could consider sponsoring a benchmark dataset.

The fact that an LLM can be used by both reporting companies and evaluators can lead to an overall improvement in the quality and impact of the disclosures. At the same time, it can also lead to an arms race where greenwashing companies expend non-productive energy in using an LLM to try to outsmart an LLMJ scoring system. Cognizance of this competing dynamic must drive all future work on this important topic.

\section*{Acknowledgments}
G.C. acknowledges funding support by a U.S. National Science Foundation Graduate Research Fellowship. %The authors are grateful to the anonymous reviewers whose comments have helped improve the paper.

%Bibliography
\bibliographystyle{unsrt}  
\bibliography{references}  

\begin{thebibliography}{10}

\bibitem{zheng2023judging}
Lianmin Zheng, Wei-Lin Chiang, Ying Sheng, Siyuan Zhuang, Zhanghao Wu, Yonghao Zhuang, Zi~Lin, Zhuohan Li, Dacheng Li, Eric Xing, et~al.
\newblock Judging llm-as-a-judge with mt-bench and chatbot arena.
\newblock {\em Advances in Neural Information Processing Systems}, 36:46595--46623, 2023.

\bibitem{stede2021climate}
Manfred Stede and Ronny Patz.
\newblock The climate change debate and natural language processing.
\newblock In {\em Proceedings of the 1st Workshop on NLP for Positive Impact}, pages 8--18, 2021.

\bibitem{rolnick2022tackling}
David Rolnick, Priya~L Donti, Lynn~H Kaack, Kelly Kochanski, Alexandre Lacoste, Kris Sankaran, Andrew~Slavin Ross, Nikola Milojevic-Dupont, Natasha Jaques, Anna Waldman-Brown, et~al.
\newblock Tackling climate change with machine learning.
\newblock {\em ACM Computing Surveys (CSUR)}, 55(2):1--96, 2022.

\bibitem{leippold2023environmental}
Markus Leippold, Dominik Stammbach, Nicolas Webersinke, Julia~Anna Bingler, and Mathias Kraus.
\newblock Environmental claim detection.
\newblock In {\em Proceedings of the 61st Annual Meeting of the Association for Computational Linguistics}, pages 1051--1066. Association for Computational Linguistics, 2023.

\bibitem{luo2020detecting}
Yiwei Luo, Dallas Card, and Dan Jurafsky.
\newblock Detecting stance in media on global warming.
\newblock {\em arXiv preprint arXiv:2010.15149}, 2020.

\bibitem{coan2021computer}
Travis~G Coan, Constantine Boussalis, John Cook, and Mirjam~O Nanko.
\newblock Computer-assisted classification of contrarian claims about climate change.
\newblock {\em Scientific reports}, 11(1):22320, 2021.

\bibitem{piskorski2022exploring}
Jakub Piskorski, Nikolaos Nikolaidis, Nicolas Stefanovitch, Bonka Kotseva, Irene Vianini, Sopho Kharazi, Jens~P Linge, et~al.
\newblock Exploring data augmentation for classification of climate change denial: Preliminary study.
\newblock In {\em Text2Story@ ECIR}, pages 97--109, 2022.

\bibitem{gehring2023analyzing}
Kai Gehring and Matteo Grigoletto.
\newblock Analyzing climate change policy narratives with the character-role narrative framework.
\newblock 2023.

\bibitem{diggelmann2020climate}
Thomas Diggelmann, Jordan Boyd-Graber, Jannis Bulian, Massimiliano Ciaramita, and Markus Leippold.
\newblock Climate-fever: A dataset for verification of real-world climate claims.
\newblock {\em arXiv preprint arXiv:2012.00614}, 2020.

\bibitem{stammbach2022environmental}
Dominik Stammbach, Nicolas Webersinke, Julia~Anna Bingler, Mathias Kraus, and Markus Leippold.
\newblock Environmental claim detection.
\newblock {\em arXiv preprint arXiv:2209.00507}, 2022.

\bibitem{morio2023nlp}
Gaku Morio and Christopher~D Manning.
\newblock An nlp benchmark dataset for assessing corporate climate policy engagement.
\newblock {\em Advances in Neural Information Processing Systems}, 36:39678--39702, 2023.

\bibitem{yim2023meticulously}
Tik~Yu Yim, Yuxuan Zhang, Wenting Tan, Tak-Wah Lam, and Siu~Ming Yiu.
\newblock Meticulously analyzing esg disclosure: A data-driven approach.
\newblock In {\em 2023 IEEE International Conference on Big Data (BigData)}, pages 2884--2889. IEEE, 2023.

\bibitem{brie2024mandatory}
Bjarne Bri{\'e}, Kristof Stouthuysen, and Tim Verdonck.
\newblock Mandatory csr reporting in europe: A textual analysis of firms’ climate disclosure narratives.
\newblock {\em Available at SSRN 4231567}, 2024.

\bibitem{webersinke2021climatebert}
Nicolas Webersinke, Mathias Kraus, Julia~Anna Bingler, and Markus Leippold.
\newblock Climatebert: A pretrained language model for climate-related text.
\newblock {\em arXiv preprint arXiv:2110.12010}, 2021.

\bibitem{vaghefi2023chatclimate}
Saeid~Ashraf Vaghefi, Dominik Stammbach, Veruska Muccione, Julia Bingler, Jingwei Ni, Mathias Kraus, Simon Allen, Chiara Colesanti-Senni, Tobias Wekhof, Tobias Schimanski, et~al.
\newblock Chatclimate: Grounding conversational ai in climate science.
\newblock {\em Communications Earth \& Environment}, 4(1):480, 2023.

\bibitem{bulian2023assessing}
Jannis Bulian, Mike~S Sch{\"a}fer, Afra Amini, Heidi Lam, Massimiliano Ciaramita, Ben Gaiarin, Michelle~Chen Huebscher, Christian Buck, Niels Mede, Markus Leippold, et~al.
\newblock Assessing large language models on climate information.
\newblock {\em arXiv preprint arXiv:2310.02932}, 2023.

\bibitem{callaghan2021machine}
Max Callaghan, Carl-Friedrich Schleussner, Shruti Nath, Quentin Lejeune, Thomas~R Knutson, Markus Reichstein, Gerrit Hansen, Emily Theokritoff, Marina Andrijevic, Robert~J Brecha, et~al.
\newblock Machine-learning-based evidence and attribution mapping of 100,000 climate impact studies.
\newblock {\em Nature climate change}, 11(11):966--972, 2021.

\bibitem{planas2022beyond}
Jordi Planas, Daniel Firebanks-Quevedo, Galina Naydenova, Ramansh Sharma, Cristina Taylor, Kathleen Buckingham, and Rong Fang.
\newblock Beyond modeling: Nlp pipeline for efficient environmental policy analysis.
\newblock {\em arXiv preprint arXiv:2201.07105}, 2022.

\bibitem{schimanski2023climatebert}
Tobias Schimanski, Julia Bingler, Camilla Hyslop, Mathias Kraus, and Markus Leippold.
\newblock Climatebert-netzero: Detecting and assessing net zero and reduction targets.
\newblock {\em arXiv preprint arXiv:2310.08096}, 2023.

\bibitem{wrzalik2024netzerofacts}
Marco Wrzalik, Florian Faust, Simon Sieber, and Adrian Ulges.
\newblock Netzerofacts: Two-stage emission information extraction from company reports.
\newblock In {\em Proceedings of the Joint Workshop of the 7th Financial Technology and Natural Language Processing, the 5th Knowledge Discovery from Unstructured Data in Financial Services, and the 4th Workshop on Economics and Natural Language Processing@ LREC-COLING 2024}, pages 70--84, 2024.

\bibitem{garigliotti2024sdg}
Dar{\'\i}o Garigliotti.
\newblock Sdg target detection in environmental reports using retrieval-augmented generation with llms.
\newblock In {\em Proceedings of the 1st Workshop on Natural Language Processing Meets Climate Change (ClimateNLP 2024)}, pages 241--250, 2024.

\bibitem{ni2023chatreport}
Jingwei Ni, Julia Bingler, Chiara Colesanti-Senni, Mathias Kraus, Glen Gostlow, Tobias Schimanski, Dominik Stammbach, Saeid~Ashraf Vaghefi, Qian Wang, Nicolas Webersinke, et~al.
\newblock Chatreport: Democratizing sustainability disclosure analysis through llm-based tools.
\newblock {\em arXiv preprint arXiv:2307.15770}, 2023.

\bibitem{colesanti2024combining}
Chiara Colesanti~Senni, Tobias Schimanski, Julia Bingler, Jingwei Ni, and Markus Leippold.
\newblock Combining ai and domain expertise to assess corporate climate transition disclosures.
\newblock {\em Available at SSRN 4826207}, 2024.

\bibitem{moodaley2023greenwashing}
Wayne Moodaley and Arnesh Telukdarie.
\newblock Greenwashing, sustainability reporting, and artificial intelligence: A systematic literature review.
\newblock {\em Sustainability}, 15(2):1481, 2023.

\bibitem{de2024will}
Charl De~Villiers, Ruth Dimes, and Matteo Molinari.
\newblock How will ai text generation and processing impact sustainability reporting? critical analysis, a conceptual framework and avenues for future research.
\newblock {\em Sustainability Accounting, Management and Policy Journal}, 15(1):96--118, 2024.

\bibitem{bingler2022cheap}
Julia~Anna Bingler, Mathias Kraus, Markus Leippold, and Nicolas Webersinke.
\newblock Cheap talk and cherry-picking: What climatebert has to say on corporate climate risk disclosures.
\newblock {\em Finance Research Letters}, 47:102776, 2022.

\bibitem{luo2024unmasking}
Yunfang Luo, Tao Yang, Qingan Li, Qiang Liu, and Xiling Cui.
\newblock Unmasking esg exaggerations using generative artificial intelligence.
\newblock 2024.

\bibitem{yue2023disc}
Shengbin Yue, Wei Chen, Siyuan Wang, Bingxuan Li, Chenchen Shen, Shujun Liu, Yuxuan Zhou, Yao Xiao, Song Yun, Xuanjing Huang, et~al.
\newblock Disc-lawllm: Fine-tuning large language models for intelligent legal services.
\newblock {\em arXiv preprint arXiv:2309.11325}, 2023.

\bibitem{son2024krx}
Guijin Son, Hyunjun Jeon, Chami Hwang, and Hanearl Jung.
\newblock Krx bench: Automating financial benchmark creation via large language models.
\newblock In {\em Proceedings of the Joint Workshop of the 7th Financial Technology and Natural Language Processing, the 5th Knowledge Discovery from Unstructured Data in Financial Services, and the 4th Workshop on Economics and Natural Language Processing@ LREC-COLING 2024}, pages 10--20, 2024.

\bibitem{xie2024doclens}
Yiqing Xie, Sheng Zhang, Hao Cheng, Pengfei Liu, Zelalem Gero, Cliff Wong, Tristan Naumann, Hoifung Poon, and Carolyn Rose.
\newblock Doclens: Multi-aspect fine-grained medical text evaluation.
\newblock In {\em Proceedings of the 62nd Annual Meeting of the Association for Computational Linguistics (Volume 1: Long Papers)}, pages 649--679, 2024.

\bibitem{chiang2024large}
Cheng-Han Chiang, Wei-Chih Chen, Chun-Yi Kuan, Chienchou Yang, and Hung-yi Lee.
\newblock Large language model as an assignment evaluator: Insights, feedback, and challenges in a 1000+ student course.
\newblock {\em arXiv preprint arXiv:2407.05216}, 2024.

\bibitem{wang2024automated}
Chihang Wang, Yuxin Dong, Zhenhong Zhang, Ruotong Wang, Shuo Wang, and Jiajing Chen.
\newblock Automated genre-aware article scoring and feedback using large language models.
\newblock {\em arXiv preprint arXiv:2410.14165}, 2024.

\bibitem{brown2020language}
Tom Brown, Benjamin Mann, Nick Ryder, Melanie Subbiah, Jared~D Kaplan, Prafulla Dhariwal, Arvind Neelakantan, Pranav Shyam, Girish Sastry, Amanda Askell, et~al.
\newblock Language models are few-shot learners.
\newblock {\em Advances in neural information processing systems}, 33:1877--1901, 2020.

\bibitem{Roucher2025using}
Aymeric Roucher.
\newblock Using llm-as-a-judge for an automated and versatile evaluation.
\newblock \url{https://huggingface.co/learn/cookbook/en/llm_judge}, 2025.
\newblock Accessed: 2025-01-16.

\bibitem{wei2022chain}
Jason Wei, Xuezhi Wang, Dale Schuurmans, Maarten Bosma, Fei Xia, Ed~Chi, Quoc~V Le, Denny Zhou, et~al.
\newblock Chain-of-thought prompting elicits reasoning in large language models.
\newblock {\em Advances in neural information processing systems}, 35:24824--24837, 2022.

\bibitem{CDP2022Scoring}
CDP.
\newblock Cdp climate change 2022 scoring methodology.
\newblock \url{https://guidance.cdp.net/en/guidance?cid=30&ctype=theme&idtype=ThemeID&otype=ScoringMethodology}, 2022.
\newblock Accessed: 2025-01-16.

\end{thebibliography}

\appendix

\newpage

\section{Probabilistic Weighting}
\label{appendix:logprobs}
The naive way to compute a score for a given response would be to simply take the outputted token of the LLM as the score. However, since LLMs sample their next tokens from a distribution of outcomes, this naive approach can be noisy, especially for small sample sizes (one sample for numerical rating, for example). Instead, we take a weighted average of the possible next tokens, weighted by the probability of sampling that token.

For any given prompt, OpenAI makes these probabilities (called ``logprobs'', because they are computed in log space) available via their API. We find that using logprobs improves the separation overall between A-List and non-A-List companies, as shown in Table~\ref{tbl:logprobs}.

\begin{table}
  \caption{Weighting by lobprobs improves separation between score distributions of A-List and non-A-List responses.}
  \centering 
  \begin{tabular}{lccc}
    \toprule
    Scoring System & TVD & KS & EMD\\
    \midrule
    %One-shot, sampled output only & 0.4289 & 0.4289 & 0.1780\\
    %One-shot, weighted by logprob & 0.4413 & 0.4423 & 0.1799\\
    Sampled output & 0.3874 & 0.3874 & 0.1802\\
    Logprob-weighted & 0.4413 & 0.4422 & 0.1842\\
  \bottomrule
\end{tabular}
\label{tbl:logprobs}
\end{table}

\newpage

% ugly minipage nonsense to force the figures to be on this page 
%\begin{minipage}{\textwidth}
\begin{figure*}
    \centering
    \begin{subfigure}[t]{0.32\textwidth} 
    \vskip 0pt
        \centering
        \includegraphics[width=\linewidth]{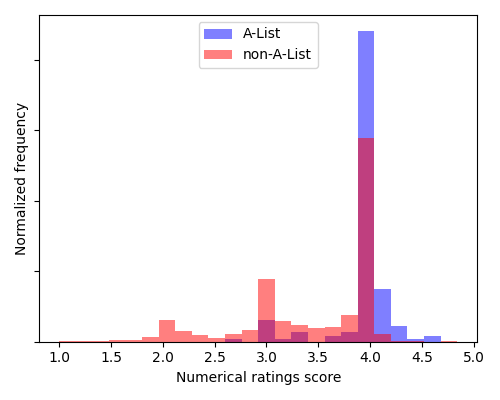}
        \caption{Zero-shot prompting}
    \end{subfigure}
    \hfill
    \begin{subfigure}[t]{0.32\textwidth} 
        \vskip 0pt
        \centering
        \includegraphics[width=\linewidth]{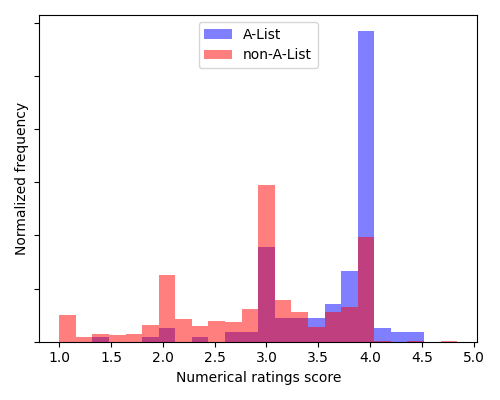}
        \caption{One-shot prompting}
    \end{subfigure}
    \hfill
    \begin{subfigure}[t]{0.32\textwidth} 
        \vskip 0pt
        \centering
        \includegraphics[width=\linewidth]{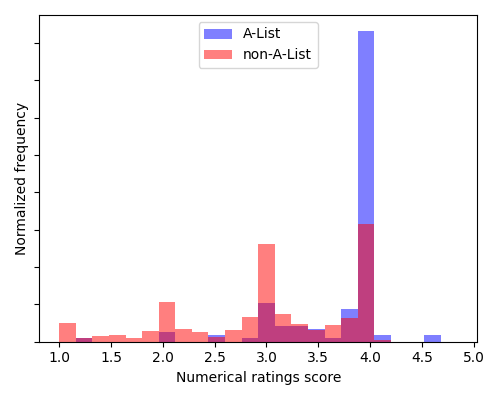}
        \caption{Two-shot prompting}
    \end{subfigure}
    \hfill
    \begin{subfigure}[t]{0.32\textwidth}
        \vskip 0pt
        \centering 
        \includegraphics[width=\linewidth]{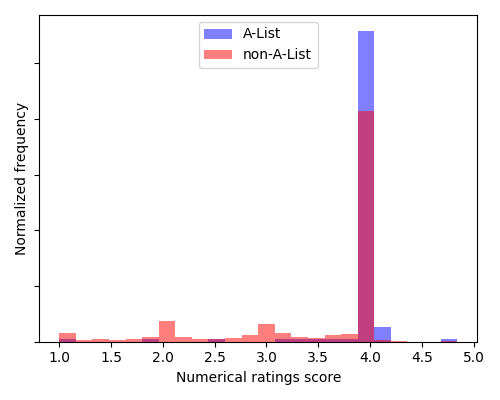}
        \caption{Zero-shot with indicative scale}
    \end{subfigure}
    \hfill
    \begin{subfigure}[t]{0.32\textwidth}    
        \vskip 0pt
        \centering 
        \includegraphics[width=\linewidth]{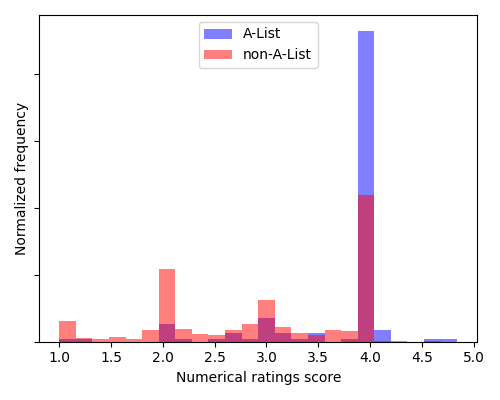}
        \caption{One-shot with indicative scale}
    \end{subfigure}
    \hfill
    \begin{subfigure}[t]{0.32\textwidth}
        \vskip 0pt
        \centering 
        \includegraphics[width=\linewidth]{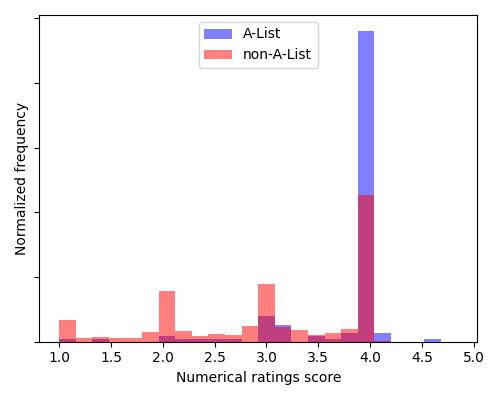}
        \caption{Two-shot with indicative scale}
    \end{subfigure}
    \caption{Distribution of numerical rating scores for various prompt configurations.}
    \label{fig:all_hists_ratings}
\end{figure*}

\begin{figure*}
    \centering
    \begin{subfigure}[t]{0.32\textwidth}
        \vskip 0pt
        \includegraphics[width=\textwidth]{fig/pairwise_distribution.png}
        \caption{No chain-of-thought prompting}
    \end{subfigure}
    \hspace{2em}
    \begin{subfigure}[t]{0.32\textwidth}
        \vskip 0pt
        \includegraphics[width=\textwidth]{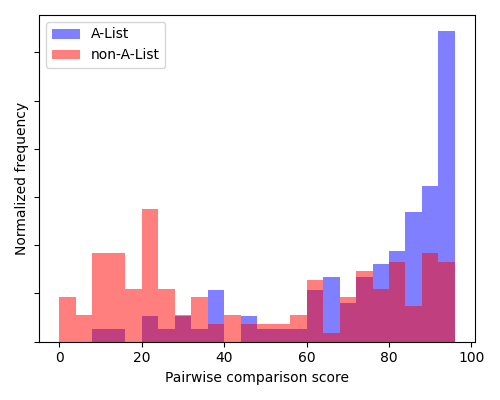}
        \caption{With chain-of-thought prompting}    
    \end{subfigure}
    \caption{Distribution of pairwise comparison scores for various prompt configurations.}
    \label{fig:all_hists_pairwise}
\end{figure*}
%\end{minipage}

\section{Score Distributions for Prompt Variants}
\label{appendix:all_hists}
We show the score distributions for A-List and non-A-List responses under each prompt variant in Figure~\ref{fig:all_hists_ratings} for the numerical rating system and Figure~\ref{fig:all_hists_pairwise} for the pairwise comparison system.

% \section{Greenwashing Prompt}
% \label{appendix:greenwashing_prompt}
% The prompt we use to generate greenwashed responses is shown in Figure~\ref{fig:greenwash_prompt}. 

% \begin{figure*}
%    \centering
%    \includegraphics[width=\linewidth]{fig/greenwash_prompt.jpg}
%    \caption{The prompt used to generate greenwashed responses. We generated responses with no constraints, accuracy constraints (red), and both accuracy (red) and length (blue) constraints.}
%    \label{fig:greenwash_prompt}
% \end{figure*}

% \section{Greenwashed Example Response}
% \label{appendix:greenwashing_example}
% A sample greenwashed response under each set of constraints is shown in Figure~\ref{fig:greenwash_example}. 

% \begin{figure*}
%    \centering
%    \includegraphics[width=\linewidth]{fig/greenwash_example.png}
%    \caption{An example CDP response, along with LLM-greenwashed variations under three sets of constraints. Changes are loosely labeled by type.}
%    \label{fig:greenwash_example}
% \end{figure*}

\section{Further Notes on LLMJ Robustness against Greenwashing}
\label{appendix:greenwashing_robustness}

We present a few additional observations about comparing the numerical rating and pairwise comparison scoring systems against LLM-greenwashed responses. 

\begin{figure}
 \centering
   \includegraphics[width=0.49\linewidth]{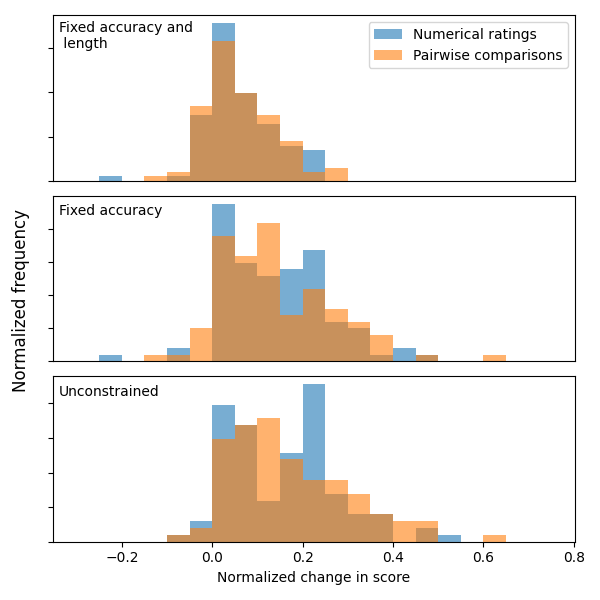}
   \caption{The two scoring systems show similar score increases under greenwashing. Numerical ratings concentrate around whole-number increases. Looser constraints result in higher score increases. }
   \label{fig:score_change_by_set}
\end{figure}

\textbf{The two scoring systems have similar distributions of overall score increases.} Figure~\ref{fig:score_change_by_set} shows the distribution of normalized change in score for each set of greenwashed responses. Overall, the distributions are very similar, with comparable peaks and tails. Notably, the tendency of the numerical ratings system to give near-integral values results in two notable peaks at 0 and 1/4 in both the “fixed accuracy” and “fixed accuracy and length” cases (whereas the pairwise distribution only has one peak). This likely contributes to the overall larger average score increase of the numerical ratings system: a score that one might “expect” to get a 0.6-point boost might instead get rounded up to 1.

\begin{figure}
   \centering
   \includegraphics[width=0.49\linewidth]{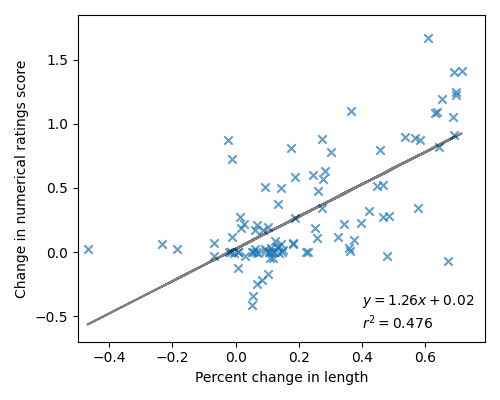}
   \caption{Responses that were lengthened more during greenwashing tended to see a larger increase in their scores.}
   \label{fig:score_vs_length}
\end{figure}

The difference between the “fixed accuracy and length” and “fixed accuracy” sets is notable. In principle, the length should have little bearing on the score of the response, especially when controlling for accuracy. However, there is a fairly noticeable jump in score increases between the two sets. This is probably due to a combination of two factors which are hard to disentangle: (a) many of the so-called “fixed accuracy” greenwashed responses have inaccurate, falsified information, and (b) \textbf{the LLM-as-a-Judge system has some association between longer responses and higher scores} (even if the extra text contains only ``fluff''). This correlation can be seen in Figure~\ref{fig:score_vs_length}: on average, a response received 0.125 more points for each 10\% increase in length.

\begin{figure}[t]
   \centering
       \begin{subfigure}[t]{0.48\textwidth}
       \centering
       \includegraphics[width=\textwidth]{fig/correlation_between_pairwise_and_ratings.png}
       \caption{On the original responses, the two scoring systems produce highly-correlated results ($r^2 = 0.70$).}
       \label{fig:pairwise_ratings_correlation_2}
   \end{subfigure}
   \hfill
   \begin{subfigure}[t]{0.48\textwidth}
       \centering
       \includegraphics[width=\textwidth]{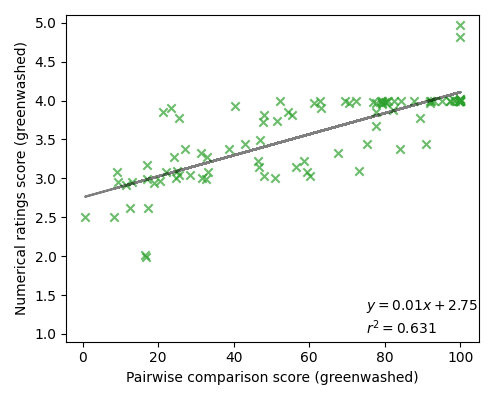}
       \caption{However, on the greenwashed responses, the scores are substantially less correlated ($r^2 = 0.63$).}
       \label{fig:greenwashed_correlation}
   \end{subfigure}
   \caption{Greenwashed responses receive less-correlated scores from pairwise comparisons and numerical ratings than the original responses.}
    \label{fig:compare_greenwashed_original_correlation}
\end{figure}

Finally, while the ratings and pairwise scores were quite correlated on the original un-greenwashed dataset (see Figure~\ref{fig:pairwise_ratings_correlation}, duplicated as Figure\ref{fig:pairwise_ratings_correlation_2} for reference), they are substantially less correlated on the greenwashed responses (Figure~\ref{fig:greenwashed_correlation}). In part, this is because the greenwashed scores are much more compressed into the numerical rating range of 3-4, while remaining quite “spread out” on the pairwise scale. However, we did not find any systematic patterns around which greenwashed responses scored very highly on one system but very poorly on the other. We speculatively note that this phenomenon seems related to the idea of Goodhart’s Law (“When a measure becomes a target, it ceases to be a good measure”): optimizing towards some scoring system (that is, greenwashing) makes the responses much more noisy on that same scoring system, rendering it less useful. 

\section{Controlling for Length}
\label{appendix:doubling_length}

Our evaluation of our scoring systems against A-List and non-A-List sets of company responses is a purely observational study: that is, we do not directly measure causal effects of the content of the response on the score. Instead, we merely establish correlation between high-scoring responses and presence on the CDP A-List. One reasonable objection might be that the LLMJ system picks up only on some superficial trait(s) of the responses (e.g., length, or some other lexical attribute) that are highly correlated with being on the A-List without truly contributing to it. For example, it is possible that good responses tend to be lengthy (and hence companies with long responses tend to be on the A-List), and that the LLMJ system is merely scoring the responses based on length rather than content. 

In Appendix~\ref{appendix:greenwashing_robustness}, we observed that the ``fixed accuracy'' set of greenwashed responses received substantially higher scores than the ``fixed length and accuracy'' set. This could be an indication that the LLMJ systems are being misled by the mere length of the response (rather than the content involved in the extra length). 

\begin{figure}
   \centering
   \includegraphics[width=0.49\linewidth]{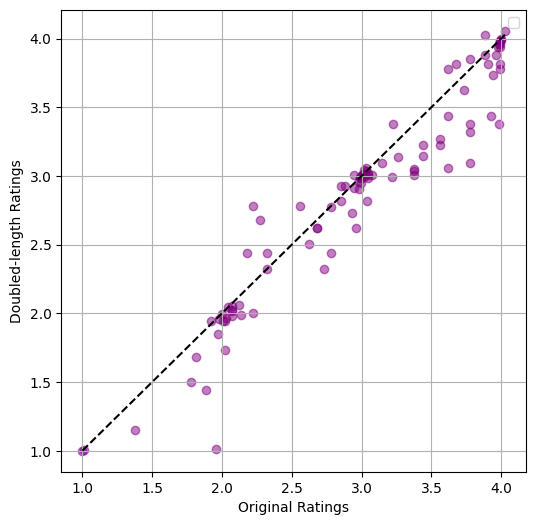}
   \caption{Original and lengthened company responses containing the same content received similar scores.}
   \label{fig:original_vs_2x_length2}
\end{figure}

To address the possibility of length being a confounding factor, we run a simple experiment in which we control for content while varying length. We use the same uniformly sampled set of 100 non-A-list companies from Section 5 and double the length of the companies' responses by repeating the response twice. We then use the numerical rating system (with one-shot learning) to score these new responses. We compare the original ratings to the new ratings in Figure~\ref{fig:original_vs_2x_length2}. We see that nearly all scores are at or below the $y=x$ line, and in fact most of the points are below the line, indicating that doubling the length \emph{reduced} the score. 

This strongly suggests that length is \emph{not} a confounding factor, and that the increase in score of the non-length-constrained greenwashed responses was due to changes in the actual content of the response rather than length alone.

\end{document}